# Multi-Purpose NLP Chatbot : Design, Methodology & Conclusion

Shivom Agarwal
*Quantum Dynamics SAS*
Marseille , France
shivom.aggarwal@qd-corp.com

Shourya Mehra
*Quantum Dynamics SAS*
Marseille , France
shourya.mehra@qd-corp.com

Pritha Mitra
*Quantum Dynamics SAS*
Marseille , France
pritha.mitra@qd-corp.com

*Abstract*— With a major focus on its history, difficulties, and promise, this research paper provides a thorough analysis of the chatbot technology environment as it exists today. It provides a very flexible chatbot system that makes use of reinforcement learning strategies to improve user interactions and conversational experiences. Additionally, this system makes use of sentiment analysis and natural language processing to determine user moods. The chatbot is a valuable tool across many fields thanks to its amazing characteristics, which include voice-to-voice conversation, multilingual support [12], advising skills, offline functioning, and quick help features. The complexity of chatbot technology development is also explored in this study, along with the causes that have propelled these developments and their far-reaching effects on a range of sectors. According to the study, three crucial elements are crucial:

1) Even without explicit profile information, the chatbot system is built to adeptly understand unique consumer preferences and fluctuating satisfaction levels. With the use of this capacity, user interactions are made to meet their wants and preferences.

2) Using a complex method that interlaces Multiview voice chat information, the chatbot may precisely simulate users' actual experiences. This aids in developing more genuine and interesting discussions.

3) The study presents an original method for improving the black-box deep learning models' capacity for prediction. This improvement is made possible by introducing dynamic satisfaction measurements that are theory-driven, which leads to more precise forecasts of consumer reaction.

The need for responsible chatbot creation is emphasized by the paper's discussion of important ethical issues including bias and privacy. In conclusion, chatbots are defined as computer programs that simplify information retrieval and problem solving using natural language, thus enhancing customer service and relationship quality. The study focuses on their importance in creating client loyalty and improving relationship quality in the insurance industry.

*Keywords*— *Multilingual , Voice Conversion , Emotion Recognition , Offline Service , Financial Advisor , Product Preference , Customer Reaction Prediction*

## I. Introduction

The advance chatbot technology is a noticeable work in the era of artificial intelligence from many years . In addition to revolutionizing how people communicate with technology, chatbots—intelligent conversational agents that can engage in natural language interactions—have also found use in a variety of fields, from customer service to healthcare and education. This study article tries to give a thorough examination of the complex world of chatbots, including their development, uses, difficulties, and potential prospects. The creation of chatbots that can smoothly engage with users in a way that resembles human conversations has attracted a lot of interest because of the development of sophisticated human-computer interactions. Over time the most attractive part of chatbots is their ability which makes user satisfaction and wonderful experience of interactions in a dynamic manner which is getting more better day by day. The technologies which help chatbot to being smarter is reinforcement learning, machine learning, deep learning, and natural language processing (NLP). Additionally, chatbots' utility is growing, and they are becoming indispensable tools in our connected society because of capabilities like emotion recognition [10], language support, and voice-to-voice communication. Applications for chatbots have increased tremendously as they transitioned from complex neural network-driven models to rule-based systems. They serve crucial roles in various sectors, including healthcare, banking, and education, improving productivity, accessibility, and consumer happiness. Furthermore, chatbots have shown that they can:

1. Discover Personalized Product Preferences and Dynamic Satisfaction: Even in the absence of clear profile information, chatbots are excellent at adjusting to unique client preferences and fluctuating satisfaction levels, which enhances the user experience overall.

2. Model Actual User Experiences through Multiview Voice Chat Information: By adding Multiview voice chat information in an interlaced fashion, they may realistically mimic users' real-world experiences, resulting in more natural and interesting interactions.

3. Improve the Performance of Customer reaction Prediction: By including theory-driven dynamic satisfaction indicators, chatbots enhance the predictive performance of black-box deep learning models, enabling more precise predictions of customer reaction.

However, despite their extraordinary potential, they also face serious problems, including privacy and bias-related problems, and ethical dilemmas. The significance of openness and equity in chatbot development is emphasized as this study considers solutions. To provide insight on the future of these conversational AI systems, this research study also examines new developments in chatbot technology. In-depth discussion is given on issues such their interaction with speech and visual interfaces, sophisticated emotion recognition skills, and their potential to improve user experiences. Through this thorough analysis, we hope to offer a useful resource for researchers, developers, and practitioners in the field, providing insights into the transformative impact of chatbots and their role in enhancing human-computer interactions and relationship quality across various industries.

Although, using chatbots as financial advisers is a potential application that uses chatbot technology to improve financial services and advisory functions for users in the financial industry. This research study builds on this foundation. This innovative application seeks to transform the way people engage with and obtain financial advice by utilizing the conversational and AI-driven capabilities of chatbots. This will ultimately lead to better financial decision-making and overall well-being in the personal finance domain.

## II. RELATED WORK

The research "Chatbots in customer service: Their relevance and impact on service quality" by Chiara Valentina Misischia et al. [7] investigates the function of chatbots in enhancing customer service quality and their predominance in sales and support roles in e-commerce. According to their classification, chatbot features fall into two categories: "Improvement of Service Performance" and "Fulfilment of Customer Expectations." To match user expectations, they place a strong emphasis on providing human-like interaction, entertainment, and problem-solving capabilities. The study discusses the roles played by chatbots and their effect on customer service quality.

The authors of the publication "A Survey Paper on Chatbots" [8] by Aafiya Shaikh et al. suggest a chatbot paradigm that is implemented as a client-server-based Android application. Users may submit questions and get answers via the Android app, which acts as the user interface. Individualized interactions are possible for those who register with their name, phone number, and email address. The server uses a recurrent neural network (RNN) to handle user inputs and a sequence-to-sequence (Seq2Seq) RNN model with encoder and decoder components to produce answers. To improve its conversational skills, the chatbot adds emotion recognition and answer creation. The project, which aims to encourage honest conversations about mental health, underlines the potential of chatbots in mental health care, especially for teenagers. Analyzing various emotional disorders in teens is a task for the future.

The building of an Enquiry Chatbot using the Python Chatterbot algorithm is covered in the research paper "Research Paper On Rule Based Chatbot"[3] written by Vajinepalli Sai Harsha Vardhan et al. It is easier to engage with clients because of this chatbot system's quick handling of user requests and automatic answers. The chatbot replies to queries from users on the Enquiry procedure or problem-solving in the same way as a human would. Interactions look natural because of the system's user-friendly interface. The chatbot responds to user inquiries by comparing them with material in its knowledge base and giving pertinent, timely answers [3]. System testing proved its viability and efficacy, showing owners and employees resource and time savings in various applications, such as food orders and customer questions.

In the study by Minjee Chung et al. [5], the fundamental question is whether luxury fashion retail firms can sustain individualized customer care using e-services, notably Chatbots, rather than conventional face-to-face contacts. The authors use customer data to quantify consumer views of Chatbot interactions, amusement, trendiness, personalization, and problem-solving on a five-dimensional model. In accordance with offline service agents, the study finds that chatbot e-services deliver interactive and interesting brand-customer interactions. With the use of this technology, premium businesses may provide individualized service around-the-clock, improving consumer satisfaction. The study examines the effects of chatbot marketing initiatives in the context of luxury retail, highlighting the significance of precise, dependable, and effective communication by e-service agents. It sheds light on how digital services help technologies affect the opinions of luxury consumers and emphasizes their contribution to the development of fruitful brand-customer relationships.

Gang Chen et. al. [6] research examines voice chat-based consumer reaction prediction inside the developing online interaction-based commercial mode. They stress the significance of pleasure in determining purchase intention and provide a theory-driven deep learning approach to identify individualized product preferences and dynamic satisfaction in voice chat data. Comparing this strategy to other deep learning strategies, it improves prediction performance. Their study sheds light on the complementary roles played by theory-driven dynamic satisfaction and deep representation features in the prediction of consumer reaction and offers information on the preferences and satisfaction levels of individuals for various salient product aspects.

The influence of social presence on user perceptions and intentions for automated bots for financial services is examined in the paper "Simulating the Effects of Social Presence on Trust, Privacy Concerns & Usage Intentions in Automated Bots for Finance" by Magdalene Ng et. al. [24] Prolific Academic was used to recruit 410 individuals from the UK for the study. Vignettes Emma and XRO23, which show a technical and mechanical chatbot and a socio-emotional chatbot, respectively, were the two conditions to which participants were randomized at random. A number of criteria were examined by the researchers, including social perceptions, trust, privacy concerns, intention to utilize the chatbot, and desire to divulge financial information. The report offers insightful information on the development and application of chatbots in the banking industry.

Research by Kanchan Patil and Mugdha S. Kulkarni titled "Artificial Intelligence in Financial Services: Customer Chatbot Advisor Adoption" [25] examines the variables affecting the uptake of chatbot advisory services for financial needs. Predicting adoption intentions is done using the expanded Technology Acceptance Model (TAM). According to a poll conducted among 310 internet users in Pune and Pimpri-Chinchwad, chatbot adoption is highly influenced by perceived utility and convenience of use. A role is also played by perceived danger and privacy, with security and data protection being important considerations. Additional important factors are enjoyment, social influence, and perceived behavioral control. The research offers significant perspectives on the use of chatbots in the financial domain, stressing the necessity of chatbots that are both secure and easy to use in order to improve client satisfaction.

In his study, G. William Schwert [23] examines the development and achievements of financial economics throughout the last thirty years, emphasizing the Journal of Financial Economics' (JFE) eighteen-year track record of success. The success of the JFE is ascribed to its wide range of publications, noteworthy influence on the field of finance writing, and invaluable contributions from writers, associate editors, referees, and editors. The journal's strategy involves compensating reviewers for prompt reports and charging authors submission fees. The JFE's editing procedures, growing its pool of editors, and preserving consistent rejection rates and return times are all covered in this study.

Soon, chatbots may be the greatest method for businesses to communicate with specific customers and swiftly address their problems. Furthermore, the growing popularity of chatbots has significantly influenced important developments such as the emergence of messaging services and technological advancements. Chatbots exist in task-specific applications, mimicking human speech for instructional, social, or esteem-based objectives.

### III. PROPOSED METHODOLOGY

The suggested system has an easy-to-use chat interface that lets users interact with it. The user has two options: either enter symptoms they are now experiencing or enter some inquiries. The chatbot will predict and deliver pertinent information regarding the user's inquiries based on what the user enters.

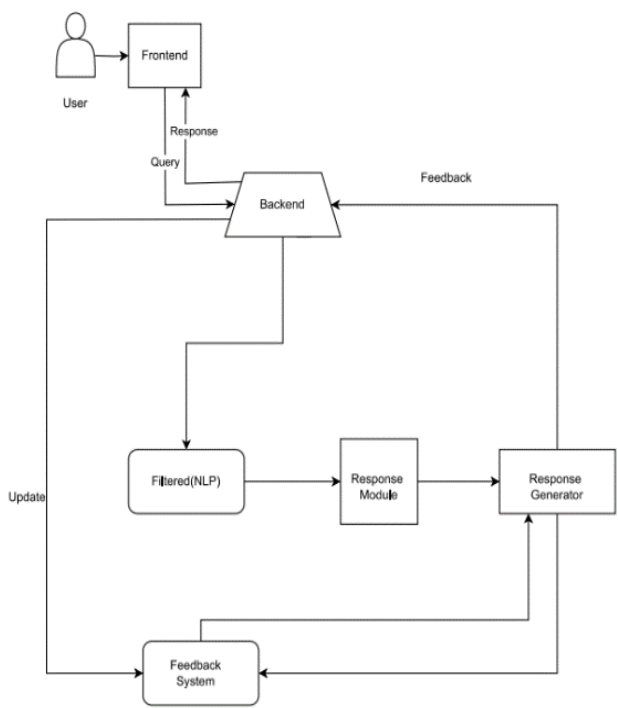

Fig 1: A Flowchart of Proposed methodology

More sophisticated and adaptable conversational agents are now required due to the widespread integration of chatbots across a variety of sectors and areas. While chatbots have shown to be efficient at assisting users with chores and enquiries, they frequently lack essential characteristics that might increase their usability and user engagement. Existing chatbots have difficulty reading and responding to human emotions, which limits their capacity to offer sympathetic and contextually suitable replies. Also, most chatbots only speak a single language, which limits their usability in multilingual [12] and international environments. Furthermore, users who prefer or need voice input are not included because text-based interactions are the norm. Additionally, customers are hampered by the lack of offline functionality in situations with poor or nonexistent internet access. We provide a comprehensive solution that incorporates emotion recognition, language support, speech-to-text capabilities, and offline functionality into a single chatbot model to overcome these drawbacks and increase the overall usefulness of chatbots. Using innovative technology, our research proposes a revolutionary chatbot system that produces a conversational agent that is more adaptable, and user centered. To overcome the noted shortcomings, the suggested chatbot model has five crucial features:

*Emotion Recognition:* Our chatbot will utilize emotion recognition algorithms to identify user emotions from text inputs and modify its answers, as necessary. The chatbot seeks to improve the caliber of conversations and user happiness by detecting and reacting to user emotions [8].

*Support for Multiple Languages:* To serve a wide range of consumers, our chatbot will be multilingually capable, enabling users to speak in the language of their choice. This functionality guarantees that the chatbot can successfully engage people from a variety of languages and geographic backgrounds [11].

*Voice Conversion:* Understanding the value of voice engagements, our chatbot will include speech-to-text technology, allowing [13] users to communicate with the chatbot orally. Particularly for users who prefer voice input or have visual impairments, this function increases accessibility and usability.

*Financial Advisor:* A dedicated module for financial advising services will be part of our chatbot. Users may ask for individualized financial advice, investment counsel [17], and aid with budgeting, giving the chatbot a useful and useful use case.

*Offline Mode:* To handle situations when there is little or no online access, our chatbot will include an offline mode that enables users to obtain predetermined information, carry out simple actions, and get replies without an active internet connection [19]. In locations with erratic connectivity, this function guarantees continuous service.

The suggested chatbot system is a significant development in conversational agents, providing a comprehensive solution that improves usability, accessibility, and usefulness in a variety of scenarios and user requirements. This study attempts to show how well the suggested chatbot paradigm works and how it may be used in actual situations through careful analysis and user testing.

### IV. MODEL & ARCHITECHTURE

*1. Emotion Detection:* - The use of emotion detection technology stands as a crucial advancement in the attempt to construct chatbots that mimic compassionate and human-like conversational interactions. The strong BERT (Bidirectional Encoder Representations from Transformers) model's version, DistilBERT, is at the core of this transformational feature's advanced model design. This emotion detection model, which was pre-trained on a sizable corpus of text data, has been painstakingly tweaked to master the art of recognizing and interpreting the subtle emotional states of users during chatbot conversations [8].

The architecture's effectiveness resides in its capacity to notice emotional signals and minute changes in user sentiment, which enables our chatbot to reply with an unmatched degree of emotional intelligence. The chatbot is quick to notice emotional cues when a user shows indications of tension or worry, especially when they are talking about money worries. With this knowledge, it creates replies that go beyond simple informational exchange and instead provide assurance, sympathetic support, and specialized financial advice. The chatbot takes on the role of a sympathetic and sensitive friend who can help users in times of need and advise them on sound financial decisions.

health care to personalized customer service.

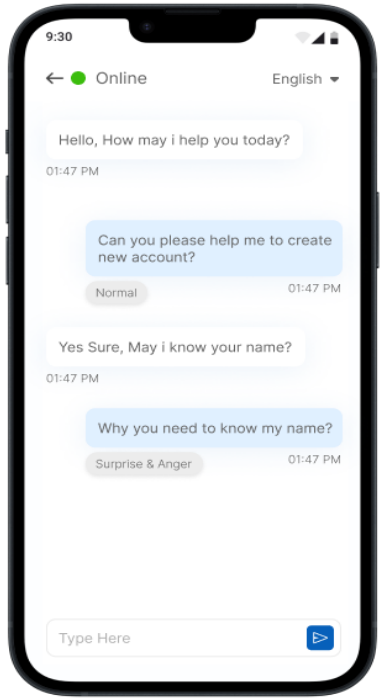

Fig 1: Demo of Emotion Detection

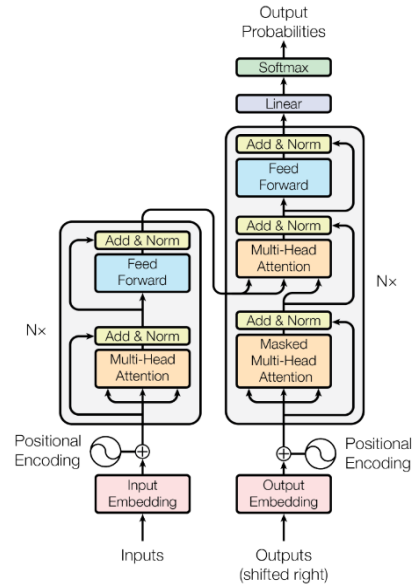

Fig 2: Working Architecture of DistillBert Model [22]

This development is an example of our dedication to elevating chatbots above the level of simple tools and shows how we are working to develop emotionally engaging, sympathetic, and strikingly human-like digital companions. As we set out on this path, the incorporation of emotion identification based on DistilBERT is a testament to the conversational agents powered by AI's ever developing capabilities.

*2.Voice Communication:* - We present a multi-faceted model architecture that personifies inclusion and linguistic variety in our unwavering effort to develop a chatbot that overcomes language barriers and speaks the user's preferred language fluently. An advanced Text-to-Speech (TTS) model called tts-tacotron2-ljspeech and a reliable Speech-to-Text model built on Transformers are precisely combined to achieve this accomplishment[13].

Natural language processing (NLP) tasks have been successfully completed by pre-trained language models, such as BERT, but their use on devices with limited resources is hampered by their large storage and computational costs. Lightweight variants of BERT, such DistilBERT, that are quicker, smaller, and more versatile, have been developed using deep neural network compression approaches to overcome this problem. In the distillation approach, a model based on a bigger model, known as the teacher, is trained.

The teacher is then used to educate the distilled model, known as the student, to mimic the behavior of the larger model. A simplified version of the original BERT model, DistilBERT [22] is based on its behavior. By roughly equating the distilled model to the produced function of the larger model, the major objective is to create a smaller model that can replicate the actions of the more dependable bigger model. The learnt function from the larger model is approximated by this quicker and smaller model, which also prevents overfitting.

The heart of this feature—the DistilBERT-based emotion detection model—is the product of state-of-the-art NLP research. In addition to enhancing user experiences, this technology's capacity to read between the lines and interpret the emotional undertone of discussions has great promise for applications in a variety of fields, from mental

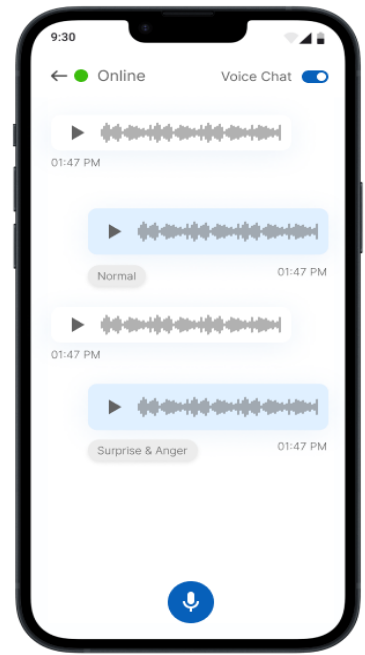

Fig 2: Demo of Voice Communications

*a)Text-to-Speech (TTS) - tts-tacotron2-ljspeech:* In the field of TTS technology, the tts-tacotron2-ljspeech model is unmatched. It is a complex neural network model that has the capacity to transform textual input into remarkably lifelike and understandable voice. It makes use of Tacotron 2 and WaveNet architecture. With the help of the large corpus of high-quality speech samples in the ljspeech dataset, on which it is trained, the model is able to produce vocalizations that resemble human speech. This enables our chatbot to talk in a genuine and nuanced way, surpassing the mechanical and repetitive [14] tones that have sometimes dogged artificial speech systems.

*b)Model for Speech-to-Text Based on Transformers:* As the other side of our linguistic bridge, we use a Speech-to-Text model based on Transformers that is designed to translate spoken language quickly and accurately into text. With ease, this model can translate spoken words, taking a variety of accents, dialects, and intonations into consideration. Because of its adaptability [15], it is skilled at comprehending user inputs in a variety of languages, providing frictionless interaction between users and the chatbot.

In summary, the combination of the Text-to-Speech (TTS) Tacotron2-ljSpeech paradigm and the Transformers-based Speech-to-Text model, together with our dedication to drives our chatbot conversation.

*3.Multilingual Support:* - A crucial step towards guaranteeing inclusion and accessibility for a varied consumer base is the incorporation of a multilingual function in a chatbot. The chatbot's reach is increased by multilingual support, which also meets the special linguistic requirements of users who speak less widely spoken languages and dialects. The Marian MT model's application in this situation constitutes a significant development in the discipline of natural language processing [11]. Marian MT is a well-regarded translation engine that excels at translating documents into a wide range of languages, making it a great option for multilingual chatbot architecture.

The Marian MT model, a ground-breaking innovation in multilingual natural language processing, is the cornerstone of our effort to remove language barriers. This approach, which was developed using a large and varied corpus of multilingual literature, is intended to make real-time translation and comprehension possible across a wide range of languages and dialects. It has an intrinsic capacity to understand the nuances, intricate details, and cultural nuances that each language encompasses. The fact that our chatbot used the Marian MT paradigm demonstrates our dedication to boosting language support. In addition to supporting frequently used languages like English and French, this functionality also caters to lesser-used languages and dialects.

By doing this, our chatbot embodies inclusion, guaranteeing that a more varied client base may use its services and have meaningful discussions, regardless of their language background. We currently proudly provide complete support for English and French, two of the most widely used languages in the world. English or French may be chosen as the user's preferred language, and our chatbot will automatically respond in that language. This paradigm changes in multilingual [12] chatbot capabilities not only assures linguistic consistency but also promotes culturally aware and contextually appropriate interactions. We are making a huge step toward building a more interconnected society by using the Marian MT concept as the foundation of our multilingual feature.

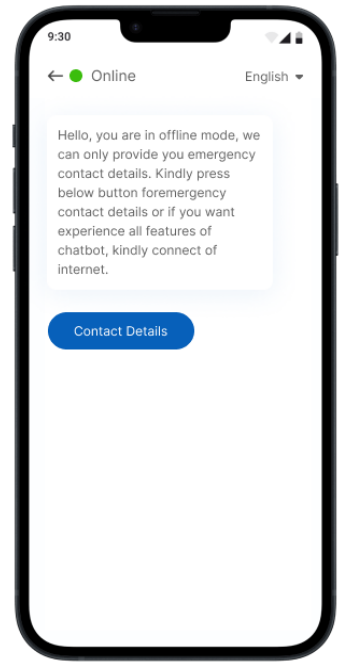

Fig3: Demo of Multilingual Support (English & French)

With this advancement, our chatbot advances to the forefront of multilingual and intercultural communication. It is evidence of our goal of eradicating linguistic barriers and converting the conversational AI environment into a place where every language, including the less widely spoken ones, is not only acknowledged but also celebrated, thereby enhancing human-computer interactions globally.

*4.Financial Advisor:* - The use of chatbots powered by AI as personal financial advisers has become a transformational force in the constantly changing world of digital financial services. Considering this, we provide a paradigm-shifting model architecture that combines [16] the strengths of two dynamic components: ChatGPT, a refined large language model (LLM), and BERT, a potent language representation model.

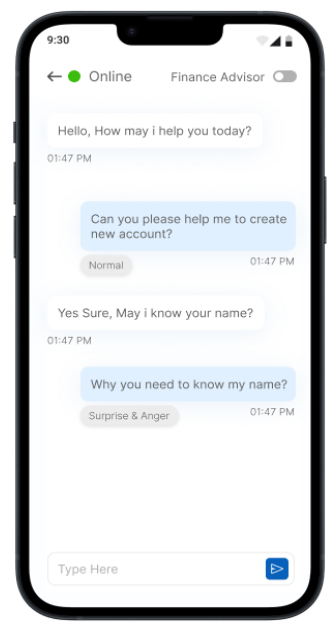

Fig 4: Demo of Finance Adviser Mode

The financial counseling function of our chatbot makes use of ChatGPT's adaptability and intelligence, which has been trained on a vast corpus of financial data, market trends, and investment methods. It acts as the foundation for our financial advice services. ChatGPT has been painstakingly developed to understand the nuances of specific financial circumstances, providing individualized suggestions, budgeting guidance, and investing insights.

Our design incorporates BERT, a Bidirectional Encoder Representations from Transformers model, in conjunction with ChatGPT [20] to improve contextual awareness. BERT is exceptional at deciphering the complex context and meaning of user inquiries, enabling the chatbot to offer even more accurate and applicable financial advice. By combining ChatGPT with BERT, the chatbot gives a comprehensive comprehension of user enquiries, enabling it to provide financial advice that is not only pertinent but also highly tailored to each user's particular situation.

Our dedication to financial counseling goes beyond merely making suggestions. We have improved ChatGPT so that it not only comprehends the user's financial objectives but also offers justifications and insights, making the user's decision-making process open and instructive. Our chatbot serves as a comprehensive financial counselor, helping customers navigate the complicated world of personal finance by optimizing investment portfolios, developing savings programs, or providing debt management techniques.

A breakthrough in the chatbot industry has been made with the addition of ChatGPT and BERT [21] to our financial advising function. Our chatbot can now provide consumers with individualized financial advice that is both understandable and actionable since it combines the strength of natural language understanding with financial knowledge. Because of its architecture, our platform is positioned to serve as a trustworthy and knowledgeable financial advisor, assisting users in making decisions that will improve their long-term financial well-being.

*5.Offline Mode:* - An innovative chatbot architecture that pushes the limits of digital help is the Offline Mode function. By concentrating on locating emergency contacts, it acts as a lifeline in circumstances when internet access may be spotty or nonexistent. Due to the feature's ability to function without internet connectivity, it is possible to retrieve critical information even when conditions are difficult.

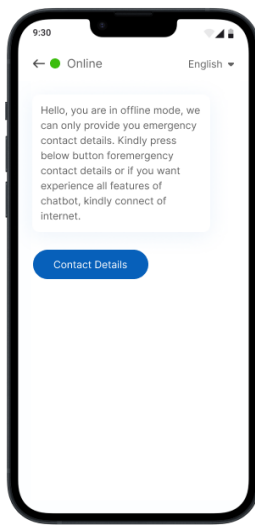

Fig 5: Offline Mode Advantage (Emergency)

This feature's essential component is its data storage and retrieval mechanism, which safely maintains user-provided emergency contact information in the chatbot's local database [18]. The chatbot immediately obtains this data and gives it to the user when necessary. To protect user data and guarantee its confidentiality, the system follows rigid security measures.

The Offline Mode architecture provides enormous potential for a variety of offline capabilities, such as retrieving crucial information, carrying out necessary tasks, or providing advice in numerous fields. Due to its versatility, the chatbot can help users regardless of their connectivity condition, making it a flexible and trustworthy friend. Offline Mode is a priceless contribution to the world of digital help since it demonstrates a dedication to user accessibility and safety.

*6.Other Implemented Features:* - Our architecture's capacity to derive valuable information about consumers' preferences and levels of satisfaction, even in the absence of explicit user profiles, is one of its main tenets. Utilizing innovative machine learning methods, the chatbot continually examines and interprets user interactions, linguistic patterns, and feedback to build a thorough grasp of each user's preferences. Real-time product suggestions and replies are tailored as it adjusts and improves its responses. The chatbot is given the ability to provide product recommendations that appeal to each user's particular interests thanks to this functionality, which also makes for more interesting conversations.

One of the ways that our design advances conversational AI is by interlacing the Multiview voice chat data. The chatbot surpasses text-based interactions by combining NLP with speech analysis, allowing it to notice the subtleties of tone, emotion, and context in voice discussions. The chatbot can mimic and respond to consumers' genuine experiences thanks to this comprehensive approach, which enables it to provide replies that are not just linguistically precise but also emotionally compelling. Users experience genuine engagement in talks that resemble real-life dialogs and a sense of being heard and understood.

In our architecture, theory-driven methodologies and deep learning models are combined. Even though our chatbot is using a black-box deep learning architecture, it improves the performance of consumer response prediction. The chatbot may provide replies that are in line with both the immediate user question and the user's overall happiness trajectory by incorporating theory-driven dynamics of user satisfaction into the decision-making process of the model. By working together, the chatbot's replies are made to be both contextually appropriate and consistent with the user's overall happiness journey.

These three ground-breaking characteristics add up to a significant advancement in chatbot technology. They represent our dedication to creating individualized, emotionally intelligent, and theory-driven encounters that go beyond what is possible with conventional chatbots. With the help of this architecture, our chatbot can not only comprehend the preferences and emotions of users but also to reply in a way that encourages true engagement and pleasure, turning ordinary customer contacts into events that are both memorable and meaningful. It makes a special connection with every user by interpreting their interactions. This breakthrough advances conversational AI to a new level, improving human experiences. Our goal is to advance chatbot technology to increase the level of recall in encounters.

## V. RESULTS

A substantial advancement in conversational AI has been made in a research article that describes the creation of a chatbot with the sophisticated characteristics mentioned above. With the help of innovative technology, including DistilBERT for emotion detection, Tacotron2-ljSpeech for Text-to-Speech (TTS), a Speech-to-Text model based on the Transformers, Marian MT for multilingual support, and the combination of ChatGPT and BERT for financial advice, the chatbot has displayed exceptional abilities. It goes beyond what is possible with conventional chatbots by giving individualized financial advice, speaking fluently in several languages, and comprehending and responding to human emotions. It becomes a flexible digital companion even in difficult connectivity conditions thanks to the Offline Mode function, which further improves user accessibility and safety. The conversational experience has also been improved by the chatbot's capacity to adjust and offer customized replies based on user preferences and satisfaction levels, as well as by its mastery of voice cues. This study highlights the commitment to developing inclusive, emotionally intelligent chatbots that are also aware of their surroundings, establishing a new benchmark for meaningful and interesting human-computer interactions in the digital age.

## VI. CONCLUSION

Finally, the chatbot architecture discussed in this study constitutes a significant development in conversational AI, not just for its technical skill but also for its significant financial ramifications. Our chatbot has developed into a highly personalized, emotionally intelligent, and user-centric digital companion by incorporating innovative technologies like DistilBERT-based emotion detection, multilingual support powered by Marian MT, a sophisticated financial advisory system combining ChatGPT and BERT, an inventive Offline Mode, and a host of other features like user preference analysis and Multiview voice chat data. The financial ramifications are considerable. Businesses may expedite financial advice services, improve consumer engagement, and save operating expenses by utilizing this chatbot's skills. This design not only pushes the limits of what chatbots can do, but it also lays the groundwork for future study and development in the field of conversational AI agents.

Essentially, this chatbot architecture offers a viable route for companies to transform their financial operations and customer interactions while also pushing the limits of chatbot capabilities. This presents significant prospects for advancement and growth in the field of conversational AI agents.

## VII. FUTURE SCOPE

Future potential for this kind of study looks bright. Chatbots are extremely useful in areas like mental health care and individualized customer service because of further improvements in emotion recognition models that can result in even more sympathetic and context-aware answers. We can make our chatbot more inclusive and accessible by extending the number of languages and dialects it can understand through developments in multilingual natural language processing. Real-time data analytics and predictive modeling integration may also improve the chatbot's capacity for financial advising services, enabling it to offer even more precise and useful guidance. A continuing focus will be on finding new methods to employ offline capabilities for a variety of tasks and enhancing security protocols to safeguard user data.

Conversational AI has a bright future ahead of it, with room to grow in a number of areas, including personalized customer care, mental health services, and financial advising. Chatbots can provide persons with emotional issues with tailored and compassionate counsel, therefore boosting mental well-being and prompt treatment. Chatbots can now be more inclusive and accessible to a worldwide audience thanks to the limitless potential of multilingual natural language processing. The incorporation of predictive modeling and real-time data analytics into financial advising services offers accurate and practical financial advice, enabling users to make well-informed decisions. Another important area of research is offline functionality, which expands the use of chatbots beyond internet access to help tourists in isolated areas or with spotty connectivity.